%% file: main.tex
\documentclass[10pt,twocolumn,letterpaper]{article}

\usepackage[pagenumbers]{cvpr}      
\usepackage{bbding}
\usepackage{color}
\usepackage{graphicx}
\usepackage{xcolor}
\usepackage{colortbl, booktabs}
\usepackage{multicol}
\usepackage{multirow}
\usepackage{comment}

\usepackage{marvosym}
\usepackage{mdframed}
\usepackage{listings}
\usepackage{amssymb}  
\usepackage{amsmath}
\usepackage{makecell}

\input{preamble}

\definecolor{cvprblue}{rgb}{0.21,0.49,0.74}
\usepackage[pagebackref,breaklinks,colorlinks,allcolors=cvprblue]{hyperref}

\title{GuideGround: VLM-guided Semantic Understanding and Viewpoint-aware Reasoning for 3D Visual Grounding}

\author{Yiwen Wang, Yuyang Deng, Yihao Long, Xi Zhao$^{\textsuperscript{\Letter}}$\\
Xi'an Jiaotong University    \\
{\tt\small{\{junzi521,dengyuyang,2214312032\}@stu.xjtu.edu.cn, xi.zhao@mail.xjtu.edu.cn}} \\
\textsuperscript{\Letter}\small{Corresponding Author: Xi Zhao.}\\
}

\begin{document}
\maketitle
\input{chapters/Abstract}
\input{chapters/Introduction}
\input{chapters/Related_Work}
\input{chapters/Method}
\input{chapters/Experiments}
\input{chapters/Conclusion}

{
    \small
    \bibliographystyle{ieeenat_fullname}
    \bibliography{ref}
}

\input{chapters/Appendix}

\end{document}

%% file: preamble.tex
\definecolor{basefirst}{HTML}{BEE5BE} 
\colorlet{first1}{basefirst!7}
\colorlet{first2}{basefirst!14}
\colorlet{first3}{basefirst!21}
\colorlet{first4}{basefirst!99}

\definecolor{basesecond}{HTML}{8EC5FF}  
\colorlet{second1}{basesecond!9}
\colorlet{second2}{basesecond!25}
\colorlet{second3}{basesecond!40}
\colorlet{second4}{basesecond!80}

\definecolor{basethird}{HTML}{C3B1E1}
\colorlet{third1}{basethird!15}
\colorlet{third2}{basethird!25}
\colorlet{third3}{basethird!40}
\colorlet{third4}{basethird!55}

\definecolor{basefourth}{HTML}{FFDAB9}
\colorlet{fourth1}{basefourth!15}
\colorlet{fourth2}{basefourth!25}
\colorlet{fourth3}{basefourth!40}
\colorlet{fourth4}{basefourth!55}
\colorlet{fourth5}{basefourth!70}
\colorlet{fourth6}{basefourth!85}
\colorlet{fourth7}{basefourth!100}

\definecolor{zero4}{HTML}{FFDAB9}



%% file: chapters/Abstract.tex
\begin{abstract}
3D visual grounding aims to localize the target object in a 3D scene from a natural language query, requiring both fine-grained semantic understanding and viewpoint-dependent spatial reasoning.
Existing methods typically formulate semantic understanding as an auxiliary closed-set object classification task and rely on multi-view feature aggregation for viewpoint reasoning, limiting semantic generalization and weakening viewpoint-specific evidence.
We observe that vision-language models naturally provide complementary capabilities through open-vocabulary semantic understanding and global scene perception.
Based on this insight, we propose GuideGround, a VLM-guided framework that complements rather than replaces task-specific grounding models by leveraging VLMs for semantic enhancement and viewpoint-specific hypothesis verification.
Specifically, we replace auxiliary closed-set object classification with VLM-generated object semantic descriptions to enhance semantic understanding. Meanwhile, instead of directly aggregating multi-view representations, we preserve viewpoint-specific grounding hypotheses through per-view grounding and explicitly verify them using VLMs across candidate viewpoints. 
Extensive experiments on the ReferIt3D benchmark demonstrate that GuideGround consistently outperforms previous state-of-the-art methods. 
Comprehensive ablation studies further confirm the effectiveness of both the proposed semantic understanding and viewpoint reasoning strategies.
\end{abstract}

%% file: chapters/Introduction.tex
\section{Introduction} 
\begin{figure}[th]
\centering
\includegraphics[width=\columnwidth]{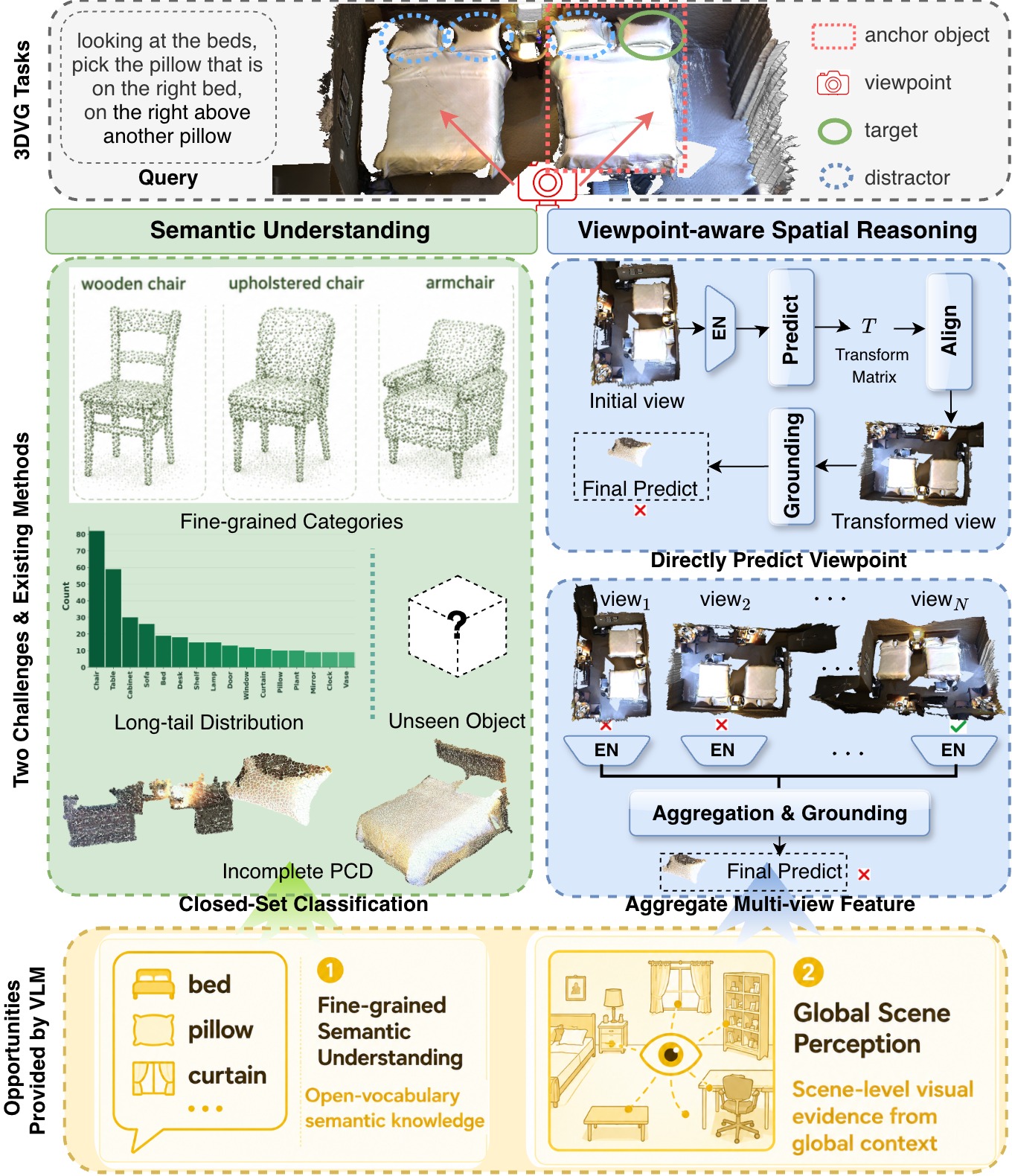} 
\caption{\textbf{Motivation of GuideGround.}
Existing 3DVG methods struggle with semantic understanding and viewpoint-aware spatial reasoning. 
}
\label{fig:teaser}
\end{figure}

3D Visual Grounding (3DVG) aims to localize a target object in a 3D scene according to a natural language description. 
As a fundamental task for 3D scene understanding, 
it serves as an essential capability for embodied AI and has attracted increasing attention in recent years~\cite{liu2025survey}.

Existing 3DVG mainly consider two settings: grounding from pre-segmented object proposals~\cite{achlioptas2020referit3d} and joint object detection and grounding~\cite{chen2020scanrefer}. 
Our work focuses on object-centric 3DVG with pre-segmented object proposals.
Despite recent progress, grounding objects in complex indoor scenes remains challenging because it depends on both accurate semantic understanding and reliable spatial reasoning.
Semantically, the model must distinguish the target object from visually or categorically similar objects. 
Spatially, it must correctly interpret viewpoint-dependent relations such as ``left'' and ``behind''. 
Existing methods improve grounding by enhancing one or both of these capabilities.

For semantic understanding, existing methods commonly employ auxiliary object classification as semantic supervision~\cite{chang2024mikasa,huang2022multi,zhang2023multi3drefer}, formulating semantic understanding as a closed-set classification problem.
This is particularly difficult in 3D indoor scenes: datasets like ScanRefer~\cite{chen2020scanrefer} and ReferIt3D~\cite{achlioptas2020referit3d} contain more than 400 fine-grained object categories, many semantically similar. In addition, they exhibit highly long-tailed distributions, unseen categories during evaluation, and incomplete object point clouds caused by occlusions.
Thus, auxiliary classification becomes a less effective semantic supervision source for 3DVG, raising a fundamental question: \textbf{Can semantic understanding in 3DVG be modeled beyond auxiliary object classification?}

For spatial reasoning, most existing methods focus on improving object-object relation modeling~\cite{yuan2021instancerefer,yang2023exploiting,chen2022language}.
However, accurate relation modeling alone cannot fully resolve viewpoint-dependent spatial reasoning. 
Existing approaches mainly follow two strategies to address this challenge. 
One strategy directly predicts the viewpoint~\cite{guo2023viewrefer,shi2024aware}. However, such predictions are ambiguous, as multiple viewpoints may satisfy the same language query. 
The other strategy samples a set of candidate viewpoints and aggregates representations from multiple rotated views~\cite{chang2024mikasa,huang2022multi}. 
While more robust, feature aggregation may blur viewpoint-specific evidence, 
leading to incorrect grounding even when one candidate viewpoint already contains sufficient evidence for identifying the target.
This raises another fundamental question: 
\textbf{Can grounding results from different viewpoints be effectively leveraged to improve viewpoint-dependent spatial reasoning?}

These findings suggest that effective 3DVG requires capabilities beyond conventional task-specific models. 
As illustrated in Figure~\ref{fig:teaser}, recent advances in vision-language models provide new opportunities for 3DVG with fine-grained semantic understanding and global scene perception~\cite{zhang2025think}. 
Rather than treating VLMs as end-to-end grounding models~\cite{li2025seeground,yang2024llm,lin2025seqvlm}, we argue that they should complement task-specific 3DVG models: VLMs provide rich semantic knowledge and scene-level visual evidence, while task-specific models remain responsible for structured spatial reasoning.

Guided by this insight, we propose \textbf{GuideGround}, a VLM-guided framework that enhances semantic understanding and viewpoint-aware spatial reasoning by complementing task-specific 3DVG models with VLMs.
First, a VLM-guided semantic understanding module generates descriptive object captions and parses the query into target and anchor descriptions, establishing semantic priors for subsequent grounding.
For viewpoint-aware spatial reasoning, an anchor-aware grounding module performs grounding independently across candidate viewpoints, preserving a grounding hypothesis for each viewpoint.
Then these hypotheses are verified by a VLM using scene-level visual observations, identifying the most plausible viewpoint for final grounding.

Our contributions are summarized as follows:
\begin{itemize}
\item 
We propose \textbf{GuideGround}, a VLM-guided framework that complements task-specific 3DVG models with vision-language models.

\item
We propose a VLM-guided semantic understanding module that enriches object representations with descriptive semantic information and parses language queries into target and anchor descriptions.

\item
We propose a VLM-guided viewpoint reasoning strategy that performs anchor-aware grounding independently across candidate viewpoints and identifies the final target through hypothesis verification.

\item
Extensive experiments on ReferIt3D demonstrate that GuideGrounder consistently outperforms previous state-of-the-art methods. 
The ablation studies further validate the effectiveness of its individual components.
\end{itemize}

%% file: chapters/Related_Work.tex
\section{Related Work}
\begin{figure*}[t]
\centering
\includegraphics[width=\textwidth]{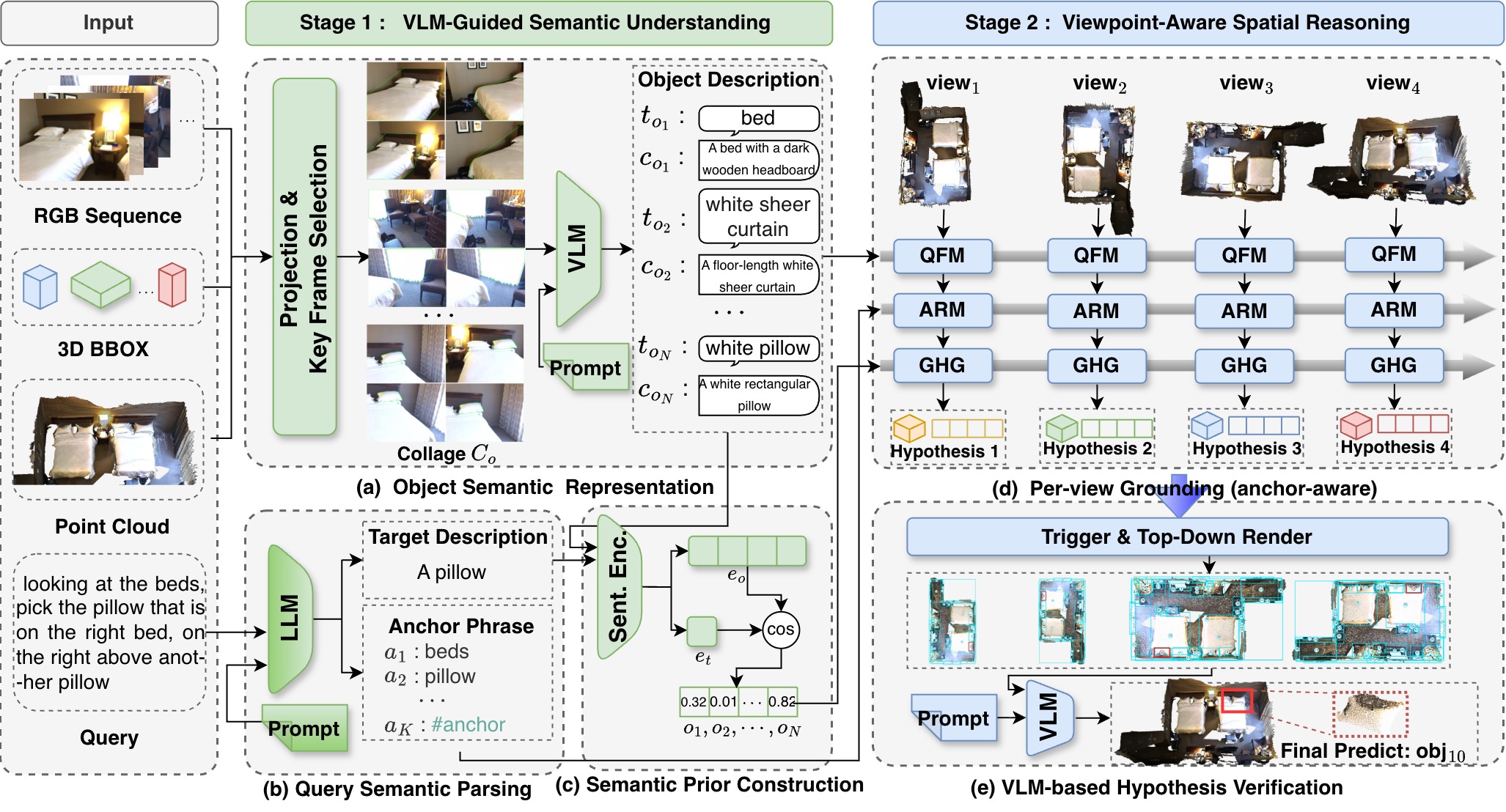} 
\caption{\textbf{Overview of GuideGround.} 
Given a 3D scene and a language query, GuideGround first performs VLM-guided semantic understanding to generate object descriptions, parse the query, and establish semantic priors. It then performs per-view grounding independently for each candidate viewpoint to produce viewpoint-specific grounding hypotheses, followed by VLM-based hypothesis verification to identify the most plausible viewpoint and determine the final grounding result.}
\label{fig:overview}
\end{figure*}

\subsection{Semantic Understanding for 3DVG}
Existing object-centric 3DVG methods improve semantic understanding by exploiting semantic information from different sources.
Most methods obtain semantic information from object point clouds through auxiliary object classification supervised by category labels
~\cite{achlioptas2020referit3d,wu2023eda,he2021transrefer3d,chen2022language,hsu2023ns3d,huang2022multi,chang2024mikasa,abdelrahman2024cot3dref}.
Another line of work introduces complementary semantic information from 2D images by exploiting pretrained vision-language models such as CLIP~\cite{radford2021learning} and DINOv2~\cite{oquab2023dinov2,jose2024dinov2meetstextunified} to obtain richer object descriptions~\cite{yang2021sat,bakr2022look,zhang2023multi3drefer}.
LanguageRefer~\cite{roh2022languagerefer} instead performs grounding in the language embedding space using object labels predicted from semantic categories.

\subsection{Spatial Reasoning for 3DVG}
Existing 3DVG methods mainly improve spatial reasoning through spatial relation modeling and viewpoint-aware reasoning.
For spatial relation modeling, early methods implicitly encode spatial relations through object coordinates or bounding-box information within graph neural networks or Transformer architectures~\cite{achlioptas2020referit3d,roh2022languagerefer,abdelrahman2024cot3dref}. Later approaches explicitly model pairwise geometric relations, such as relative distances, directions and sizes~\cite{huang2021text,yuan2021instancerefer,yang2023exploiting,chen2022language,he2021transrefer3d,hsu2023ns3d}. 
For viewpoint-aware reasoning, existing methods either predict the observer's viewpoint from the scene and language~\cite{shi2024aware} or reason over multiple candidate viewpoints~\cite{huang2022multi,chang2024mikasa,huang2025viewsrd}. 

\subsection{Vision-Language Models for 3DVG}
Existing studies mainly leverage VLMs in two paradigms: end-to-end grounding models or general vision-language representation learners.
The former either convert 3D scenes into structured semantic descriptions and perform grounding through language reasoning~\cite{yang2024llm,huang2024chat,fang2024transcrib3d,yuan2024visual}, or directly reason over RGB observations\cite{li2025seeground,liu2026view,mcvay2025locate,jain2025unifying}. 
The latter incorporates VLMs into unified vision-language learning frameworks by jointly optimizing grounding together with other 3D scene understanding tasks, such as scene captioning and visual question answering~\cite{jain2025unifying,zhu20233d}.
Our work instead exploit VLM's strengths in open-vocabulary semantic understanding and global scene perception to enhance specialized 3DVG models while preserving the structured spatial reasoning capability of task-specific grounding networks.

%% file: chapters/Method.tex
\section{Method}

\subsection{Overview}
Figure~\ref{fig:overview} presents the overall pipeline of GuideGround, which performs 3D visual grounding in two stages.
In the first stage, the semantic understanding module employs a pretrained VLM to generate descriptive object captions and parses the language query into target and anchor descriptions, establishing semantic priors for grounding.
In the second stage, the viewpoint-aware spatial reasoning module first performs anchor-aware grounding independently for each candidate viewpoint to produce viewpoint-specific grounding hypotheses. These hypotheses are subsequently verified using scene-level visual observations to identify the most plausible viewpoint and determine the final grounding result.

\subsection{VLM-Guided Semantic Understanding}
The semantic understanding module constructs semantic representations for objects and extracts structured cues from the query, providing semantic priors for subsequent grounding. 

\textbf{Object Semantic Representation.}
Given a segmented object instance $o$, we first project its associated 3D points back onto the captured RGB images to collect object-centric visual observations.
Since a single observation is often incomplete due to occlusions, multiple informative frames are selected and arranged into a unified collage $C_o$, as shown in Figure~\ref{fig:overview} (a).
The collage is then fed into a pretrained VLM $\mathcal{M}_{cap}$  to obtain $(t_o,c_o)$, where \(t_o\) is a concise object description that captures the object's primary semantic attributes, and \(c_o\) is a more comprehensive semantic description containing richer appearance details and contextual information.

\textbf{Query Semantic Parsing.}
Given a language query $Q$, we employ an LLM $\mathcal{M}_{par}$ to extract a target description $t_Q$ together with a set of anchor descriptions $\mathcal A=\{a_1,\cdots,a_K\}$. 
The target description summarizes the semantic attributes of the object to be grounded, while the anchor descriptions describe contextual objects that provide spatial cues for grounding.
The target description is used to estimate semantic correspondence, whereas the anchor descriptions are subsequently used to guide anchor-aware spatial reasoning.

\textbf{Semantic Prior Construction.}
The detailed object description $c_o$ captures richer semantic cues and is used to estimate semantic correspondence with the parsed target description $t_Q$.
Specifically, $c_o$ and $t_Q$ are encoded by the same sentence embedding model $\mathcal{M}_{sent}$ to obtain semantic embeddings $\mathbf e_o$ and $\mathbf e_t$, respectively. The semantic prior is then computed as

\begin{equation}
s_o^{\text{sem}}=
\frac{\mathbf e_o^\top\mathbf e_t}
{\|\mathbf e_o\|\|\mathbf e_t\|},
\label{eq:semantic_prior}
\end{equation}
where $s_o^{\text{sem}}$ measures the semantic consistency between candidate object $o$ and the target description $t_Q$. 

Details of the frame selection strategy and the prompts used for semantic description generation and query parsing are provided in the Supplementary Material.

\subsection{Viewpoint-aware Spatial Reasoning}
\begin{figure}[t]
\centering
\includegraphics[width=\columnwidth]{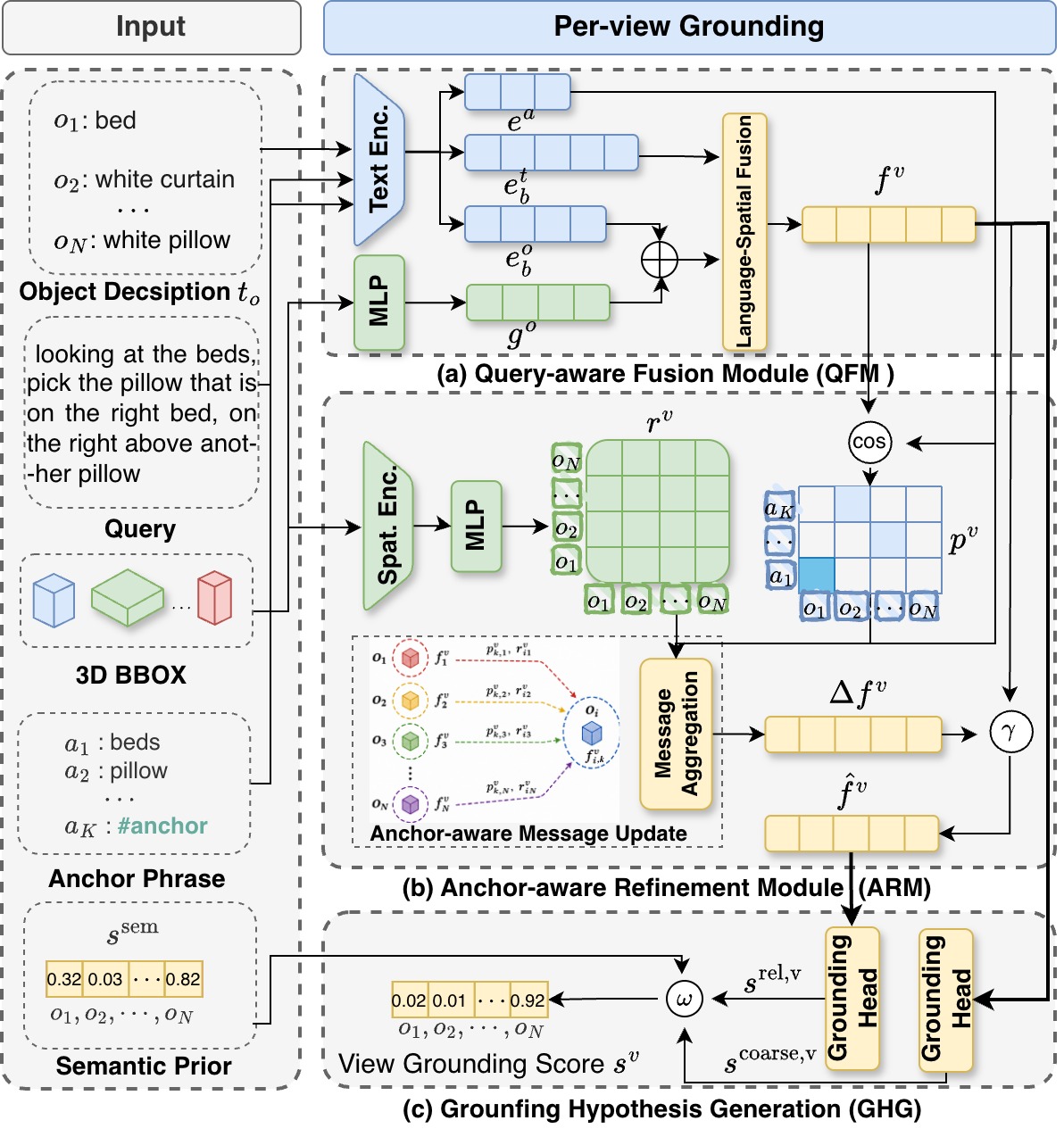} 
\caption{\textbf{Architecure of per-view grounding pipeline.} 
QFM produces query-conditioned object representations from the language query, object descriptions, and geometric information. ARM further refines them through anchor localization and anchor-aware spatial reasoning, while GHG generates the grounding hypothesis for the current viewpoint.}
\label{per_view_grounding}
\end{figure}

To resolve viewpoint ambiguity, we decompose viewpoint-aware spatial reasoning into two stages: per-view grounding and hypothesis verification. Rather than directly aggregating multi-view representations, the former generates an independent grounding hypothesis for each viewpoint, while the latter verifies these hypotheses to determine the final referent.

\textbf{Per-view Grounding.}
For clarity, all formulations in this section are presented under a single viewpoint $v$. 
As shown in Figure~\ref{per_view_grounding}, per-view grounding is accomplished through three sequential modules: Query-aware Fusion (QFM), Anchor-aware Refinement (ARM), and Grounding Hypothesis Generation (GHG).

We first employ a Query-aware Fusion Module (QFM) to fuse object semantic features, coarse geometric cues, and the language query, producing query-conditioned object representations ${\mathbf f_i^{v}}_{i=1}^{N}$.
The QFM is built upon the language-spatial fusion design of MiKASA~\cite{chang2024mikasa}, while replacing point-cloud object features extracted by PointNet++ with language embeddings of the object descriptions $t_o$ generated by the semantic understanding module.

We then refine the query-conditioned object representations using an Anchor-aware Refinement Module (ARM), which first localizes anchor objects and subsequently performs anchor-aware spatial reasoning.
For each anchor phrase $a_k$, we first estimate its correspondence with every scene object by
\begin{equation}
\ell_{k,i}^{v}
=
\frac{
(\mathbf e_k^{a})^\top \mathbf f_i^{v}
}{
\tau\,
\|\mathbf e_k^{a}\|\,
\|\mathbf f_i^{v}\|
},
\qquad
p_{k,i}^{v}
=
\operatorname*{softmax}_i
(\ell_{k,i}^{v}),
\label{eq:anchor_matching}
\end{equation}
where $\mathbf e_k^{a}$ denotes the language embedding of anchor phrase $a_k$, $p_{k,i}^{v}$ denotes the probability that object $o_i$ corresponds to anchor phrase $a_k$, and $\tau$ is the temperature parameter.
The predicted anchor distributions are then used to construct anchor-aware messages by jointly considering encoded relative geometry, object features, and anchor phrase representations according to
\begin{equation}
\mathbf m_{k,i}^{v}
=
\phi\!
\left(
\sum_{j=1}^{N}
p_{k,j}^{v}
\mathbf r_{ij}^{v}
,
\sum_{j=1}^{N}
p_{k,j}^{v}
\mathbf f_j^{v}
,
\mathbf e_k^a
\right),
\label{eq:anchor_message}
\end{equation}
where $\mathbf r_{ij}^{v}$ denotes the encoded relative geometric representation between objects $o_i$ and $o_j$ and $\phi(\cdot)$ is an MLP.
The messages from all valid anchor phrases are aggregated to obtain the refinement feature $\Delta\mathbf f_i^{v}$ according to
\begin{equation}
w_k^{v}
=
\max_{j}
p_{k,j}^{v},
\qquad
\Delta\mathbf f_i^{v}
=
\frac{
\sum_{k=1}^{K}
w_k^{v}\,
\mathbf m_{k,i}^{v}
}{
\sum_{k=1}^{K}
w_k^{v}
}.
\label{eq:message_aggregation}
\end{equation}
$\Delta\mathbf f_i^{v}$ then is injected into the query-conditioned representation through a gated residual connection as
\begin{equation}
\hat{\mathbf f}_i^{v}
=
\mathbf f_i^{v}
+
\gamma
\Delta\mathbf f_i^{v},
\label{eq:arm}
\end{equation}
where $\gamma$ is a learnable residual gate.

Finally, the Grounding Hypothesis Generation (GHG) module produces a viewpoint-specific grounding hypothesis score by integrating the semantic consistency score $s_i^{\mathrm{sem}}$, the coarse grounding score $s_i^{\mathrm{coarse},v}$ from the QFM, and the relation-aware grounding score $s_i^{\mathrm{rel},v}$ from the ARM as
\begin{equation}
s_i^{v}
=
\lambda_s
\,s_i^{\mathrm{sem}}
+
\lambda_c
\,s_i^{\mathrm{coarse},v}
+
\lambda_r
\,s_i^{\mathrm{rel},v},
\end{equation}
where $\lambda_s$, $\lambda_c$, and $\lambda_r$ are the corresponding fusion weights.
The predicted object under viewpoint $v$ is obtained by
\begin{equation}
\hat y^v
=
\arg\max_i s_i^v.
\end{equation}
The predicted object $\hat y^v$, together with its grounding score distribution, forms the grounding hypothesis for viewpoint $v$, which is subsequently examined by the hypothesis verification module to determine the final referent.

\textbf{VLM-based Hypothesis Verification.}
The per-view grounding module produces one grounding hypothesis for each viewpoint. As shown in Figure~\ref{fig:overview}(e), we further evaluate the consistency and confidence of these hypotheses, and invoke the verification VLM $\mathcal{M}_{ver}$ only when the grounding results are insufficiently reliable.
Specifically, if all viewpoint-specific predictions are identical, the prediction is directly accepted.
Otherwise, let $s_{\max}$ and $s_{\mathrm{2nd}}$ denote the highest and second-highest grounding scores after score fusion. 
If $s_{\max}-s_{\mathrm{2nd}}>\delta$, where $\delta$ denotes the confidence threshold, the prediction is also accepted. 
Only the remaining ambiguous samples proceed to the VLM verification stage.

For each viewpoint, we construct a verification hypothesis containing complementary visual and semantic evidence defined as
\begin{equation}
\mathcal{H}^{v}=
\left\{
I_{\mathrm{top}}^{v},
\hat{y}^{v},
\mathcal{P}^{v}
\right\}
\end{equation}
where $I_{\mathrm{top}}^{v}$ denotes a top-down scene visualization highlighting the predicted target object together with all query-related objects, 
 $\hat{y}^{v}$ is the viewpoint-specific grounding prediction, and $\mathcal{P}^{v}$ represents the semantic descriptions of the corresponding context objects infered during semantic understanding. 
Together, these cues provide sufficient evidence for assessing the consistency between the grounding prediction and the language query.

The generated hypotheses $\{\mathcal{H}^{v}\}_{v=1}^{V}$ and the language query $Q$ are jointly provided to the VLM $\mathcal{M}_{ver}$, which determines the most plausible referent $\hat{y}^{\mathrm{vlm}}$.

Unlike end-to-end VLM-based grounding methods, the VLM is only responsible for comparing the candidate hypotheses and selecting the one that is most consistent with the language query, while the spatial reasoning itself is performed by the task-specific grounding model.

\begin{table*}[t]
\small
\centering
\begin{tabular}{lc|ccccc|ccccc}
\toprule
\multirow{2}{*}{Method} & \multirow{2}{*}{Sem.} 
& \multicolumn{5}{c|}{Nr3D} 
& \multicolumn{5}{c}{Sr3D} \\
\cmidrule(lr){3-7} \cmidrule(lr){8-12}
&  & Overall & Easy & Hard & VD & VID 
& Overall & Easy & Hard & VD & VID \\
\midrule
UniVLG      & RGB-D   & \underline{65.2} & --   & --   & --   & --   & \textbf{81.7} & --   & --   & --   & -- \\
Locate3D$^{*}$    & RGB-D   & 56.1 & --   & --   & --   & --   & 68.2 & --   & --   & --   & -- \\
\midrule
SAT               & PCD+RGB & 49.2 & 56.3 & 42.4 & 46.9 & 50.4 & 57.9 & --   & --   & --   & -- \\
CoT3DRef          & PCD     & 64.4 & 70.0 & 59.2 & 61.9 & \underline{65.7} & 73.2 & 75.2 & 67.9 & 67.6 & 73.5 \\
CORE-3DVG$^{*}$   & PCD     & 49.6 & 54.0 & 45.4 & 46.9 & 48.2 & 50.1 & 54.3 & --   & --   & -- \\
ViL3DRel          & PCD     & 64.4 & 70.2 & 57.4 & 62.0 & 64.5 & 72.8 & 74.9 & 67.9 & 63.8 & 73.2 \\
MVT-3DVG          & PCD     & 55.1 & 61.3 & 49.1 & 54.3 & 55.4 & 64.5 & 66.9 & 58.8 & 58.4 & 64.7 \\
LanguageRefer     & PCD     & 43.9 & 51.0 & 36.6 & 41.7 & 45.0 & 56.0 & 58.9 & 49.3 & 49.2 & 56.3 \\
MiKASA            & PCD     & 64.4 & 69.7 & \underline{59.4} & \underline{65.4} & 64.0 & 75.2 & 78.6 & 67.3 & \underline{70.4} & 75.4 \\
3D-VisTA$^{*}$    & PCD     & 64.2 & \underline{72.1} & 56.7 & 61.5 & 65.1 & 76.4 & \underline{78.8} & \textbf{71.3} & 58.9 & \textbf{77.3} \\
ViewRefer         & PCD     & 56.0 & 63.0 & 49.7 & 55.1 & 56.8 & 67.0 & 68.9 & 62.1 & 52.2 & 67.7 \\
MVT-ScanEnts      & PCD     & 59.3 & 65.4 & 53.5 & 57.3 & 60.4 & --   & 
 -- & --  & -- & -- \\
 \midrule
\textbf{Ours}     & PCD + RGB        & \textbf{68.1}   & \textbf{74.6}   & \textbf{61.8}   & \textbf{66.6}   & \textbf{68.8}   & \underline{76.5}   & \textbf{78.9}   & \underline{70.9}   & \textbf{79.6}   & \underline{76.1}   \\
\bottomrule
\end{tabular}

\caption{Comparison with state-of-the-art specialized models on ReferIt3D validation split. 
``Sem.'' indicates the source of semantic information used for object representation.
VD and VID denote view-dependent and view-independent queries, respectively.
* indicates that the results are predicted from the detected bounding boxes on acc@0.25.
}
\label{tab:sota_referit3d}
\end{table*}

\subsection{Training Objective.}

Only the per-view grounding module is optimized during training, while the semantic understanding and hypothesis verification modules remain frozen.
Since each query corresponds to a single target object, directly supervising each viewpoint-specific prediction may introduce inconsistent optimization signals across viewpoints.
Therefore, we aggregate grounding scores across viewpoints during training, while preserving individual viewpoint hypotheses during inference for VLM-based verification.

Specifically, let $s_i^v$ denote the grounding score of object $o_i$ under viewpoint $v$. The aggregated score is computed as
\begin{equation}
s_i^{\mathrm{agg}}
=
\frac{1}{V}
\sum_{v=1}^{V}s_i^v ,
\end{equation}
where $V$ denotes the number of candidate viewpoints.
The grounding network is optimized with a combination of the Cross-Entropy grounding loss $\mathcal{L}_{\mathrm{ref}}$ and two auxiliary objectives:
\begin{equation}
\mathcal{L}
=
\mathcal{L}_{\mathrm{ref}}
+
\lambda_a\mathcal{L}_{\mathrm{anchor}}
+
\lambda_d\mathcal{L}_{\mathrm{dis}}.
\end{equation}
$\mathcal{L}_{\mathrm{anchor}}$ and $\mathcal{L}_{\mathrm{dis}}$ are anchor correspondence and hard-negative ranking losses, respectively. $\lambda_a$ and $\lambda_d$ are their corresponding loss weights.
$\mathcal{L}_{\mathrm{anchor}}$ encourages the model to align anchor phrases with their corresponding objects, while $\mathcal{L}_{\mathrm{dis}}$ improves discrimination between the target object and semantically similar distractors.
Detailed formulations are provided in the Supplementary Material.

%% file: chapters/Experiments.tex
\section{Experiments}

\subsection{Experimental Setup}
\textbf{Datasets and Evaluation.}
We evaluate GuideGround on the ReferIt3D benchmark~\cite{achlioptas2020referit3d}, built upon ScanNet~\cite{dai2017scannet}, under the standard pre-segmented object proposal setting.
ReferIt3D consists of two complementary subsets: Nr3D, which contains 41.5K free-form natural language descriptions, and Sr3D, which contains 83.5K template-based descriptions emphasizing fine-grained spatial relations. 
Following the official protocol, Nr3D is further divided into \emph{Easy}/\emph{Hard} and \emph{View-Dependent (VD)}/\emph{View-Independent (VID)} subsets. We report grounding accuracy on the overall test set and all evaluation subsets.
For anchor supervision, we use the entity annotations provided by ScanEnts3D~\cite{abdelreheem2024scanents3d}.

\textbf{Implementation Details.}
Our framework is implemented in PyTorch and trained on three NVIDIA RTX 4090 GPUs (32 GB). 
We use the AdamW optimizer~\cite{loshchilov2017decoupled} with a batch size of 8 for 120 epochs. During training, only the per-view grounding module is optimized, while all pretrained models remain frozen. 
DeepSeek-V3-0324~\cite{liu2024deepseek} is used for query parsing, Qwen3.5-397B-A17B~\cite{qwen3.5} for object caption generation, GPT-5.6 Luna\cite{openai2026gpt56} for hypothesis verification, and MPNet~\cite{song2020mpnet} as the sentence embedding model for semantic similarity estimation. 
Additional implementation details, prompt templates, and hyperparameter settings are provided in the Supplementary Material.

\subsection{Comparison with State-of-the-Art Methods}
We compare GuideGround with two kinds of methods: specialized 3DVG models and zero-/few-shot approaches.
To facilitate a fair comparison, we report the semantic sources and object proposal settings of different methods.

Table~\ref{tab:sota_referit3d} compares GuideGround with existing specialized 3DVG methods on ReferIt3D.
GuideGround achieves the best overall performance on Nr3D and remains highly competitive on Sr3D, demonstrating the effectiveness of combining VLM-enhanced semantic understanding with viewpoint-aware reasoning.
On Nr3D, GuideGround improves the previous best overall accuracy from 65.2\% to 68.1\%, with consistent gains on challenging Hard and VD subsets (61.8\% and 66.6\%, respectively).
These improvements indicate that VLM-enhanced semantics help distinguish fine-grained objects, while preserving viewpoint-specific hypotheses improves grounding under view-dependent descriptions.
On Sr3D, GuideGround achieves 76.5\% overall accuracy.
Although Sr3D is generated from predefined spatial templates with less semantic and viewpoint ambiguity, GuideGround still achieves strong performance on VD queries (79.6\%), suggesting that viewpoint-aware reasoning remains beneficial for spatial relation understanding.

\begin{table}[t]
\small
\centering
\begin{tabular}{l|ccccc}
\toprule
Method & \multicolumn{5}{c}{Nr3D} \\
\cmidrule(lr){2-6}
& Overall & Easy & Hard & VD & VID \\
\midrule
SeeGround         & 46.1 & 54.5 & 38.3 & 42.3 & 48.2 \\
WS-3DVG           & 39.0 & 46.5 & 31.7 & 36.8 & 40.0 \\
ZSVG3D            & 39.0 & 46.5 & 31.7 & 36.8 & 40.0 \\
VLM-Grounder      & 48.0 & 55.2 & 39.5 & 45.8 & 49.4 \\
View-on-Graph     & 47.6 & \underline{58.9} & 37.2 & 39.4 & 52.1 \\
SeqVLM            & \underline{53.2} & 58.1 & \underline{47.4} & \underline{51.0} & \underline{54.5} \\
\midrule
\textbf{Ours}     & \textbf{68.1}   & \textbf{74.6}   & \textbf{61.8}   & \textbf{66.6}   & \textbf{68.8} \\
\bottomrule
\end{tabular}
\caption{Comparison with SOTA Zero-/Few-shot methods on NR3D validation split. VD/VID definitions same as above.}
\label{tab:sota_vlm_nr3d}
\end{table}
Table~\ref{tab:sota_vlm_nr3d} compares GuideGround with recent zero-/few-shot grounding methods that directly leverage pretrained vision-language or large language models without task-specific 3D grounding training.
GuideGround achieves a substantial improvement over these approaches, increasing the overall accuracy from 53.2\% to 68.1\%.
This result indicates that pretrained foundation models are more effective when serving as complementary semantic and reasoning components within a dedicated 3D grounding framework, rather than being directly used as standalone predictors.
By combining VLM-enhanced semantic understanding with task-specific viewpoint-aware spatial reasoning, GuideGround effectively exploits the strengths of both foundation models and specialized 3D representations.

\subsection{Ablation Study}
We conduct ablation studies on NR3D to evaluate the effectiveness of each component of GuideGround.

\textbf{Does replacing auxiliary object classification with VLM-generated semantic priors improve grounding?}
We compare different semantic representations while keeping the remaining grounding framework unchanged. 
Specifically, we compare three representative semantic representations based on PointNet++, CLIP, and their combination (PN++ \& CLIP), where object semantics are learned through the same auxiliary object classification objective.
Our method instead replaces this objective with VLM-generated object descriptions and semantic priors.

Table~\ref{tab:ablation_semantic} shows that replacing auxiliary object classification with VLM-generated semantic priors consistently improves grounding performance across all evaluation subsets.
Compared with the strongest conventional semantic representation based on PointNet++, our method improves the overall grounding accuracy from 64.4\% to 68.1\%.
These results validate our motivation that semantic understanding in 3DVG can be more effectively modeled through VLM-generated semantic priors than conventional auxiliary object classification.
Moreover, simply combining PointNet++ and CLIP features does not outperform PointNet++, indicating that introducing additional visual features alone is insufficient without explicit semantic priors.

\begin{table}[ht]
\small
\centering
\begin{tabular}{lcccccc}
\toprule
Semantic Rep. & Overall & Easy & Hard & VD & VID \\
\midrule
PointNet++                     & 64.4 & 71.6 & 57.5 & 64.4 & 64.3 \\
CLIP                           &  53.3    & 59.3     & 47.5     &  53.1    &  53.3    \\
PN++ \& CLIP        &  61.5    &  69.1    &  54.2    &    60.9  &  61.8    \\
\midrule
\textbf{Ours}       & \textbf{68.1} & \textbf{74.6} & \textbf{61.8} & \textbf{66.6} & \textbf{68.8} \\
\bottomrule
\end{tabular}
\caption{
Comparison of different semantic representations. All variants use the same grounding framework.
}
\label{tab:ablation_semantic}
\end{table}

\begin{figure*}[t]
\centering
\includegraphics[width=0.95\textwidth]{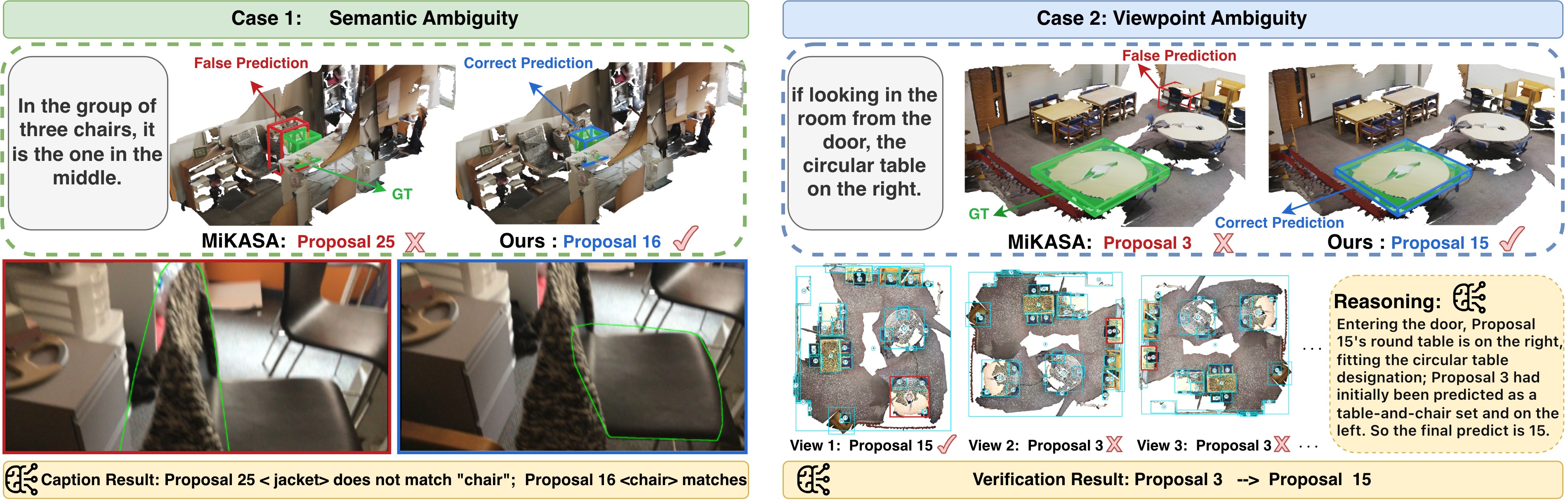}
\caption{
Qualitative examples of GuideGround.
Case 1 highlights the benefit of VLM-enhanced semantic evidence under incomplete geometric observations, while Case 2 demonstrates viewpoint-specific hypothesis verification.}
\label{fig:quality_result}
\end{figure*}

\textbf{Does VLM-guided viewpoint reasoning improve grounding performance?}
We compare different strategies for viewpoint-aware grounding while keeping the semantic representation unchanged.
Specifically, we compare conventional multi-view feature fusion, direct foundation-model reasoning (text-only and text+image), and our VLM-guided viewpoint reasoning framework. 
We further remove ARM and the VLM-based hypothesis verification module to evaluate the contribution of each component.

As shown in Table~\ref{tab:view_reasoning_ablation}, our method consistently outperforms all competing reasoning strategies.
Compared with conventional multi-view feature fusion, our framework improves the overall accuracy from 64.7\% to 68.1\%, demonstrating that preserving viewpoint-specific grounding hypotheses is more effective than aggregating multi-view features before grounding.
Direct foundation-model reasoning alone performs substantially worse, even when both textual and visual observations are provided.
For a fair comparison, the direct foundation-model baselines use the same foundation model as the verification stage, with either textual scene descriptions or both textual descriptions and rendered images as input.
This indicates that the improvement does not simply come from applying a stronger foundation model, but from integrating its semantic and visual reasoning capability with structured 3D grounding.

To isolate the contribution of each design, we additionally remove ARM and the VLM-based hypothesis verification module from the full model.
Removing ARM decreases the accuracy from 68.1\% to 63.4\%, demonstrating the importance of anchor-aware object-relation modeling for fine-grained spatial reasoning. 
Removing hypothesis verification also causes a performance drop from 68.1\% to 65.2\%, showing that multimodal verification effectively resolves ambiguous viewpoint-specific grounding hypotheses. 
Together, these results confirm that per-view grounding and VLM-based verification provide complementary benefits.
\begin{table}[ht]
\small
\centering
\begin{tabular}{lccccc}
\toprule
Strategy & Overall & Easy & Hard & VD & VID \\
\midrule
\multicolumn{6}{c}{\textit{Existing Reasoning Paradigms}}\\
\midrule
MV Fusion              & 64.7   & 71.2  & 58.4   & 63.2  & 65.5 \\
\midrule
VLM(Text-only)              & 28.6   & 36.0  & 21.5   & 26.9  & 29.5    \\
VLM(Text+Image)             & 54.6   & 58.1  & 51.3   & 49.1  & 59.4   \\
\midrule
\multicolumn{6}{c}{\textit{Component Ablation}}\\
\midrule
w/o ARM.               & 63.4   & 69.8  & 57.3   & 63.1  & 63.6   \\
w/o verification       & 65.2   & 72.4  & 58.3   & 63.7  & 66.0   \\
\midrule
\textbf{Ours}           & \textbf{68.1} & \textbf{74.6} & \textbf{61.8} & \textbf{66.6} & \textbf{68.8} \\
\bottomrule
\end{tabular}
\caption{Comparison of different spatial reasoning strategies. All variants use the same semantic representation.}
\label{tab:view_reasoning_ablation}
\end{table}

\textbf{How does the number of candidate viewpoints affect grounding?}
We analyze the influence of the number of candidate viewpoints while keeping other components unchanged.
As shown in Table~\ref{tab:view_number}, performance improves as the number of viewpoints increases from 1 to 4, indicating that additional views provide more complete scene observations.
However, using 8 viewpoints introduces redundant hypotheses and slightly degrades performance.
Therefore, we select 4 viewpoints as a good trade-off between viewpoint coverage and reasoning complexity.
\begin{table}[ht]
\small
\centering
\begin{tabular}{lccccc}
\toprule
\# Views & Overall & Easy & Hard & VD & VID \\
\midrule
1 & 60.1   & 66.5   & 53.8   & 60.0   & 61.8 \\
2 & 64.4   & 70.6   & 58.5   & 62.4   & 65.5 \\
4 & 68.1   & 74.6   & 61.8   & 66.6   & 68.8   \\
8 & 65.5   & 72.0   & 59.3   & 64.3   & 66.2 \\
\bottomrule
\end{tabular}
\caption{Effect of the number of candidate viewpoints.}
\label{tab:view_number}
\end{table}

\subsection{Qualitative Analysis}
Figure~\ref{fig:quality_result} presents two representative examples illustrating how GuideGround resolves semantic and viewpoint ambiguities.
In the first example, the baseline model grounds the query to the jacket instead of the target chair because occlusion introduces misleading geometric cues, making the jacket appear similar to a chair back.
GuideGround correctly identifies the target by leveraging VLM-enhanced semantic evidence.
In the second example, different viewpoints produce inconsistent grounding hypotheses. GuideGround preserves these hypotheses and employs VLM-based verification to select the most plausible candidate based on visual evidence and spatial relations.
These examples demonstrate that semantic enhancement and viewpoint-aware hypothesis verification address complementary challenges in 3D visual grounding.
More qualitative results are provided in the Supplementary Material.

%% file: chapters/Conclusion.tex
\section{Conclusion}
In this paper, we presented GuideGround , a VLM-guided framework for 3D visual grounding that addresses semantic understanding and viewpoint-aware spatial reasoning.
By leveraging VLM-generated semantic descriptions and viewpoint-specific hypothesis verification, GuideGround improves grounding beyond auxiliary closed-set classification and conventional multi-view featureaggregation.
Experiments on ReferIt3D demonstrate the effectiveness of the proposed framework, highlighting the potential of integrating foundation models with specialized 3D perception models for semantic-rich and reasoning-aware 3D understanding.

%% file: chapters/Appendix.tex
\appendix
\clearpage
 \setcounter{page}{1}
 \maketitlesupplementary

 \section{Additional Method Details}
\subsection{Key Frame Selection}
\label{sec:keyframe_selection}
For each object $o$, we select a small set of informative RGB frames for object description generation. To balance visual quality and viewpoint diversity, we adopt an object-aware key-frame selection strategy instead of uniform temporal sampling or naive top-$N_f$ selection.

\paragraph{Candidate frame construction.}
Let $P_o\subset\mathbb{R}^{3}$ denote the point cloud of object $o$. 
We subsample the RGB sequence using a temporal stride of $6$. 
Each sampled frame consists of an RGB image $I_t$ and the corresponding target-instance mask $M_{o,t}$. 
We project $P_o$ onto each sampled frame using the corresponding camera parameters. The frames without valid image-plane projections will be discarded.  
To reduce computation, at most $2{,}500$ object points are used when evaluating each object--frame pair.

\paragraph{Frame quality evaluation.}
For each valid object--frame pair ($o,t$), we compute three complementary quality measures. 
The projection coverage ratio is defined as
\begin{equation}
r^{\mathrm{cov}}_{o,t}
=
\frac{
  \left|
    \left\{
      \mathbf{p}\in P_o : \pi_t(\mathbf{p})\in\Omega_t
    \right\}
  \right|
}{
  |P_o|
},
\end{equation}
where $\pi_t(\cdot)$ denotes projection onto frame $t$, and $\Omega_t$ denotes its image domain. 
This ratio measures the fraction of the object geometry that falls within the camera view.

Multiple 3D points may project onto the same pixel.
We retain the front-most projected point according to depth and define the visibility ratio as
\begin{equation}
r^{\mathrm{vis}}_{o,t}
=
\frac{
    |\mathcal{U}^{\mathrm{front}}_{o,t}\cap M_{o,t}|
}{
|\mathcal{U}^{\mathrm{front}}_{o,t}|
},
\end{equation}
where $\mathcal{U}^{\mathrm{front}}_{o,t}$ is the set of pixels occupied by the front-most projections of object $o$. 
This measure estimates the fraction of the projected object that is consistent with the observed instance mask and therefore remains visible after occlusion.

We additionally compute the normalized projected area
\begin{equation}
r^{\mathrm{area}}_{o,t}
=
\frac{\operatorname{area}(B_{o,t})}{H_tW_t},
\end{equation}
where $B_{o,t}$ is the tight 2D bounding box of the valid projection and $ H_t\times W_t $ is the image resolution.
The overall frame-quality score is
\begin{equation}
\phi(o,t)
=
\lambda_{\mathrm{cov}}r^{\mathrm{cov}}_{o,t}
+
\lambda_{\mathrm{vis}}r^{\mathrm{vis}}_{o,t}.
\end{equation}
We set $\lambda_{\mathrm{cov}}=0.65$, $\lambda_{\mathrm{vis}}=0.35$. 

Although projected area is not included in the ranking score, it is retained as a hard filtering criterion to prevent very small object regions from being selected.
Specifically, a frame is considered qualified only if
\begin{equation}
r^{\mathrm{cov}}_{o,t}\geq 0.08,\qquad
r^{\mathrm{vis}}_{o,t}\geq 0.10,\qquad
r^{\mathrm{area}}_{o,t}\geq 0.002.
\end{equation}
These conditions remove frames in which the object is largely outside the camera frustum, heavily occluded, or too small to provide useful appearance information.

\paragraph{Segment-wise key-frame selection.}
We select $N_f=4$ key frames for each object. 
Since the highest-scoring frames are often temporally adjacent, directly selecting the top-$N_f$ candidates results in redundant observations. 
Instead, we sort the qualified candidates by timestamp and partition them into $N_f$ contiguous temporal segments,
\begin{equation}
\mathcal{Q}_o
=
\mathcal{Q}^{(1)}_o
\cup \cdots \cup
\mathcal{Q}^{(N_f)}_o.
\end{equation}
The highest-scoring candidate is selected from each segment:
\begin{equation}
\hat{q}_k
=
\arg\max_{q\in\mathcal{Q}^{(k)}_o}
\phi(q),
\qquad k=1,\ldots,N_f.
\end{equation}
The highest-scoring frame is selected from each segment. 
If fewer than $N_f$ qualified candidates are available, the remaining positions are filled using the highest-quality unselected candidates. Objects without any valid candidate are skipped.

\paragraph{Multi-view evidence construction.}
For each selected frame, we highlight the visible support of the target instance with a green contour. 
Compared with a rectangular 2D box, the contour more precisely identifies the target region while retaining the surrounding scene context required for recognizing object attributes. 
The four selected frames are resized and arranged into a $2\times2$ collage,
\begin{equation}
C_o
=
\Psi
\left(
I_{\hat{t}_1},
I_{\hat{t}_2},
I_{\hat{t}_3},
I_{\hat{t}_4}
\right),
\end{equation}
which is subsequently provided to the VLM for object description generation.

\subsection{Candidate Viewpoint Construction}

Given a scene coordinate system, we generate candidate viewpoints by rotating the scene around the vertical axis with uniformly sampled angles. For $V$ viewpoints, the rotation angle of the $v$-th view is defined as
\begin{equation}
\theta_v=\frac{2\pi v}{V}.
\end{equation}
The same candidate viewpoints are used for all objects within a scene. Geometric relations are recomputed under each viewpoint, while language queries and object semantic descriptions remain unchanged.

\section{Training Details}
\subsection{Training Objective}
As we describe in the main paper, viewpoint-specific grounding scores are aggregated during training to obtain a unified supervision target, whereas inference preserves the per-view grounding hypotheses for subsequent verification.
Specifically, let $s_i^{v}$ denote the final grounding score of object $o_i$ under viewpoint $v$. The aggregated training score is computed by
\begin{equation}
s_i^{\mathrm{agg}}
=
\frac{1}{N_v}
\sum_{v=1}^{N_v}
s_i^{v},
\end{equation}
where $N_v$ is the number of candidate viewpoints and we set it to 4. The aggregated prediction is supervised using the standard cross-entropy loss
\begin{equation}
\mathcal{L}_{\mathrm{ref}}
=
\mathrm{CE}
(
\mathbf{s}^{\mathrm{agg}},
y
),
\end{equation}
where $y$ denotes the ground-truth target object.

Besides the grounding loss, we employ two auxiliary objectives. The first is a multi-positive anchor loss
\begin{equation}
\mathcal{L}_{\mathrm{anchor}}
=
-\frac{1}{|\mathcal{A}^{+}|}
\sum_{k \in \mathcal{A}^{+}}
\log
\sum_{i\in\mathcal{P}_k}
\frac{\exp(\ell_{k,i})}
{\sum_j\exp(\ell_{k,j})},
\end{equation}
where $\mathcal{P}_k$ denotes the set of valid objects corresponding to anchor phrase $a_k$.
$\mathcal{L}_{\mathrm{anchor}}$ supervises anchor-object correspondence when multiple valid anchor objects exist. The second is a margin-based ranking loss which further separates the target object from hard distractors of the same category. It is defined as
\begin{equation}
\mathcal{L}_{\mathrm{dis}}
=
\frac{1}{|\mathcal{D}|}
\sum_{j\in\mathcal{D}}
\max
\left(
0,
m-s_y+s_j
\right)
\end{equation}
where $s_y$ denotes the grounding score of the target object, $s_j$ is the grounding score of distractor object $o_j$. $\mathcal D$ denotes all non-target objects sharing the same semantic category as the target. 
$m$ is the ranking margin and we set it to 0.2.

The overall objective is
\begin{equation}
\mathcal{L}
=
\mathcal{L}_{\mathrm{ref}}
+
\lambda_a
\mathcal{L}_{\mathrm{anchor}}
+
\lambda_d
\mathcal{L}_{\mathrm{dis}},
\end{equation}
where $\lambda_a$ and $\lambda_d$ are the loss weights for the auxiliary objectives. We set $\lambda_a=0.05$ and $\lambda_d=0.1$.

\subsection{Optimization Settings}
We optimize the trainable modules using AdamW with a base learning rate of $5\times10^{-4}$ and a weight decay of $0.05$. 
The query encoder, post-object encoder, and base fusion layers are optimized with a learning rate of $5\times10^{-5}$, while the newly introduced caption mapping, box mapping, relation branch, grounding classifier, and anchor refinement branch use the base learning rate. 
The learning rate is decayed by a factor of $0.66$ at epochs 30, 40, 50, 60, 70, 80, 90, 100, 110, and 120. 
We train the model for 120 epochs with a batch size of 8 on three NVIDIA RTX 4090 GPUs.

\subsection{Hyperparameter Settings}
\paragraph{Anchor-related hyperparameters.}
We extract at most $K$ anchor phrases from each referring expression. 
The anchor-object matching score is computed with a temperature parameter $\tau$ to control the sharpness of the similarity distribution. 
In our implementation, we set the maximum number of anchors to $K=6$ and use $\tau=0.1$.

\paragraph{Grounding score aggregation.}
The final grounding score combines coarse object-level matching and relation-aware refinement scores. 
Specifically, the grounding score is computed as a weighted combination of the coarse grounding score and the relation-enhanced score:
\begin{equation}
s_i^{v}
=
\lambda_s
\,s_i^{\mathrm{sem}}
+
\lambda_c
\,s_i^{\mathrm{coarse},v}
+
\lambda_r
\,s_i^{\mathrm{rel},v},
\end{equation}
where $\lambda_s$, $\lambda_c$, and $\lambda_r$ control the contributions of the three components. 
We set $\lambda_s=0.2$, $\lambda_c=1.0$, and $\lambda_r=0.2$ in all experiments.

\section{Prompt Templates}
\subsection{Object Caption Prompt}
We use the following prompts for object-centric caption generation. 
Given a \(2\times2\) multi-view collage of a target object, the VLM is instructed to generate a structured description based only on visible evidence. 
\paragraph{Primary system prompt.}
\begin{mdframed}
\small

You caption one highlighted indoor object from a 2x2 multi-view collage.
In every tile, the same target object is marked by a green outline or contour.
Use only the visible evidence in the images. Do not assume any hidden GT label.
Return JSON only.
\end{mdframed}  

\paragraph{Primary user prompt.}
\begin{mdframed}
\small
\textbf{Task:} Describe the highlighted object shown across four views.\\

\textbf{Important context:}
The collage contains 4 viewpoints of the same target object. 
The target is the region enclosed by the green outline or contour in each tile. 
Some views may be blurry, partial, or heavily occluded.
Use the combination of all views before deciding the object category.\\

\textbf{Output requirements:}\\
- Return strict JSON only.\\
- Use this exact schema:\\
\texttt{\{"target": "short object noun phrase", "caption": "one concise sentence"\}}\\
- `target` should be a short noun phrase, ideally 1 to 4 words.\\
- `caption` should be one concise sentence describing the object category and a few visible distinguishing cues.\\
- Do not mention the green outline, collage layout, or camera viewpoints in the caption.\\
- Avoid generic targets like `object` unless the object is truly unidentifiable.\\[5pt]

\textbf{Example output:}\\
\texttt{
\{"target": "toilet",
"caption": "A white ceramic toilet with the lid raised beside a tiled wall."\}
}

\end{mdframed}

The prompt design aims to ensure that (1) the VLM focuses only on the target object, (2) the generated description is grounded in visible evidence, and (3) the output follows a structured format for automatic parsing.

\subsection{Query Parsing Prompt}
We provide the prompt template used for query parsing. 
The parsing process is performed by a single LLM call, which jointly identifies the target description and reference entities required for subsequent anchor-aware reasoning. 
\paragraph{Primary system prompt.}
\begin{mdframed}
\small

You are an assistant for 3D visual grounding.
Given a referring expression describing an object in a 3D scene, extract the target object phrase and reference anchors required for spatial reasoning.
Analyze only the provided language query. Do not assume any additional scene information.
Return JSON only.

\end{mdframed}

\paragraph{Primary user prompt.}
\begin{mdframed}
\small

\textbf{Task:} Parse the referring expression for 3D visual grounding.\\

\textbf{Input:}\\
A language query referring to a target object in a 3D scene.\\

\textbf{Target object extraction:}\\
- Identify the phrase that describes the target object to be grounded.\\
- Preserve the original wording whenever possible.\\
- Include necessary semantic modifiers that define the target object.\\
- Do not include relational descriptions that only indicate the target object's location.\\

\textbf{Anchor extraction:}\\
- Identify reference objects or entities used to locate the target object.\\
- Each anchor should correspond to an object phrase or meaningful reference entity in the query.\\
- Preserve the original wording whenever possible.\\
- Do not include the target object itself as an anchor.\\
- Do not include pure spatial words (e.g., left, right, above) without an associated reference object.\\
- If no valid anchor exists, return an empty list.\\[5pt]

\textbf{Output requirements:}\\
- Return strict JSON only.\\
- Use this exact schema:\\
\texttt{
\{"target": "target object phrase", 
"anchors": ["anchor phrase 1", "anchor phrase 2"]\}
}\\[5pt]

\textbf{Example input:}\\
\texttt{
Find the mug on the right shelf next to the coffee machine.
}\\[5pt]

\textbf{Example output:}\\
\texttt{
\{"target": "the mug",
"anchors": ["the right shelf", "the coffee machine"]\}
}
\end{mdframed}
\subsection{Hypothesis Verification Prompt}
We provide the prompt template used for VLM-based grounding hypothesis verification.
The output is formatted as a structured JSON object for automatic parsing.

\paragraph{Primary system prompt.}
\begin{mdframed}
\small

You are a 3D visual grounding verifier.
Given an Nr3D referring expression, multiple top-view box images, and the corresponding textual information for each view, your task is to verify the grounding hypothesis.
The images contain proposal IDs. The textual information provides proposal ID, object ID, object category mappings, and model predictions for each view.
Use the query, object categories, and spatial layouts across different views to determine the most plausible grounding result.
Return JSON only.

\end{mdframed}

\paragraph{Primary user prompt.}
\begin{mdframed}
\small

\textbf{Task:} Verify the grounding hypothesis for the given referring expression.\\

\textbf{Important context:}\\
- The input contains a referring expression, multiple top-view images, and view-specific object information.\\
- The images contain proposal IDs corresponding to objects in the scene.\\
- The textual information provides the mapping between proposal IDs, object IDs, object descriptions, and model predictions.\\
- The grounding model has generated predictions from different views, which should be considered as strong evidence.\\
- Only override the model prediction when the visual spatial evidence clearly contradicts it.\\[5pt]

\textbf{Decision requirements:}\\
- Combine the referring expression, object categories, and spatial layouts across views to determine the most plausible grounding hypothesis.\\
- Select the view that provides the most useful evidence for interpreting the referring expression.\\
- If the expression contains viewpoint-dependent terms (e.g., left, right, front, behind, facing), select the view that best explains the corresponding directional relationship.\\
- If the expression contains spatial relations involving relative positions (e.g., closest, farthest, near, next to, between, above, below), select the view that most clearly reveals the involved objects and their relationships.\\
- If multiple views provide sufficient evidence, prefer the view where the model prediction is more stable and consistent with the object category information.\\[5pt]

\textbf{Input information:}\\
\texttt{example\_id}: <example\_id>\\
\texttt{scene\_id}: <scene\_id>\\
\texttt{ann\_id}: <ann\_id>\\
\texttt{utterance}: <utterance>\\[3pt]
\texttt{Model grounding prediction}: <fused\_prediction>\\[3pt]
\texttt{Available view IDs}: <available\_view\_ids>\\[3pt]
\texttt{View-specific textual information}: \\
\{proposal IDs, object IDs, object descriptions, and confidence scores for each view.\}\\[3pt]

\textbf{Output requirements:}\\
- Return strict JSON only.\\
- Use this exact schema:\\
\texttt{
\{"selected\_view": 0,
"selected\_proposal\_id": 0,
"selected\_object\_id": 0,
"confidence": 0.0,
"reason": "short reason"\}
}\\
- \texttt{selected\_view} must be selected from the provided view IDs.\\
- \texttt{selected\_proposal\_id} should correspond to the selected grounding hypothesis.\\
- \texttt{selected\_object\_id} should be filled from the object mapping when available; otherwise output -1.\\
- \texttt{confidence} should be a value between 0 and 1.\\
- \texttt{reason} should be a concise one-sentence explanation.\\[5pt]

\textbf{Example input:}\\
\begin{lstlisting}[basicstyle=\small\ttfamily,numbers=none,breaklines=true]
example_id: example_00000
scene_id: scene0565_00
ann_id: 0
utterance:
The chair to the far left hand side of the taller desk with the computer monitor on it.

Model grounding prediction:
proposal 30

Available view IDs:
[0,1,2,3]

View 0:

proposal -> object_description
proposal 0: grey office chair
proposal 1: office chair
proposal 2: teal office chair
proposal 3: computer monitor

Top predictions:
proposal 30: purple office chair, score=7.349187
proposal 2: teal office chair, score=4.801944
proposal 1: office chair, score=0.843593
proposal 0: grey office chair, score=0.087762
proposal 4: black desk, score=0.054076

View 1:
...
\end{lstlisting}

\textbf{Example output:}\\
\texttt{
\{"selected\_view": 2,
"selected\_proposal\_id": 12,
"selected\_object\_id": 45,
"confidence": 0.92,
"reason": "View 2 clearly shows the chair on the right side of the table."\}
}

\end{mdframed}
\subsection{VLM-based Baseline Prompt}
For the VLM-only baselines, we use simplified prompts that directly request the target object proosal from either the language query alone or the query-image pair.
In the text-only setting, the VLM receives only the referring expression and structured proposal information, including object categories and 3D bounding box coordinates. No rendered images are provided. 
This setting evaluates the capability of language-based reasoning with explicit geometric descriptions.
In the image-text setting, the VLM receives the referring expression together with rendered scene images. 
Unlike our verification module, no grounding predictions are provided. Instead, the VLM is directly asked to identify the target object from the visual input.
This setting evaluates direct VLM-based 3D visual grounding without specialized grounding models.
The exact prompts are included in the released code.
\section{Additional Experimental Results}

\subsection{Empirical Analysis of Multi-view Fusion}

\begin{table}[b]
\centering
\begin{tabular}{lccc}
\toprule
\textbf{Method} & \textbf{Fusion} & \textbf{View-Wise} & \textbf{$\Delta$ } \\
\midrule
MVT-3DVG & 54.8 & 68.9 & +14.1 \\
MiKASA   & 55.3 & 76.9 & +21.6 \\
\bottomrule
\end{tabular}
\caption{Comparison between the original multi-view fusion strategy and per-view evaluation. 
View-wise evaluation performs grounding independently on each candidate viewpoint while keeping the remaining inference pipeline unchanged.}
\label{tab:motivation}
\end{table}

\begin{table*}[t]
\centering
\small
\setlength{\tabcolsep}{3.0pt}
\begin{tabular}{lccc|ccc|ccc|ccc|ccc}
\toprule

& \multicolumn{3}{c|}{Overall}
& \multicolumn{3}{c|}{Easy}
& \multicolumn{3}{c|}{Hard}
& \multicolumn{3}{c|}{View-dep.}
& \multicolumn{3}{c}{View-indep.}
\\

\cmidrule(lr){2-4}
\cmidrule(lr){5-7}
\cmidrule(lr){8-10}
\cmidrule(lr){11-13}
\cmidrule(l){14-16}

Method
& Acc.
& S. Acc.
& D. Err.
& Acc.
& S. Acc.
& D. Err.
& Acc.
& S. Acc.
& D. Err.
& Acc.
& S. Acc.
& D. Err.
& Acc.
& S. Acc.
& D. Err.
\\

\midrule

MVT
& 54.90 & 83.50 & 28.59
& 62.44 & 79.83 & 17.39
& 47.65 & 87.02 & 39.37
& 52.93 & 84.93 & 31.99
& 55.88 & 82.79 & 26.90
\\

MiKASA
& 64.28 & 88.23 & 23.94
& 69.80 & 84.66 & 14.85
& \underline{58.98} & \underline{91.66} & 32.69
& \underline{65.43} & 89.59 & \textbf{24.16}
& 63.71 & 87.55 & 23.84
\\

Ours w/o Veri.
& \underline{65.18} & \underline{88.92} & \underline{23.74}
& \underline{72.36} & \underline{86.94} & \underline{14.58}
& 58.27 & 90.83 & \underline{32.56}
& 64.53 & \underline{90.31} & 25.79
& \underline{65.50} & \underline{88.23} & \underline{22.73}
\\

Ours + Veri.
& \textbf{68.05} & \textbf{90.05} & \textbf{21.99}
& \textbf{74.60} & \textbf{88.36} & \textbf{13.76}
& \textbf{61.76} & \textbf{91.66} & \textbf{29.91}
& \textbf{66.55} & \textbf{91.16} & \underline{24.62}
& \textbf{68.80} & \textbf{89.49} & \textbf{20.70}
\\

\bottomrule
\end{tabular}
\caption{Semantic error analysis on Nr3D across different splits.
S. Acc. (Semantic Accuracy) measures whether the predicted object belongs to the same semantic class as the target.
D. Err. (Distractor Error) denotes the gap between the S. Acc. and the final grounding performance.
Veri. denotes the VLM-based verification module.
}

\label{tab:nr3d_semantic_error}
\end{table*}

\begin{table*}[t]
\centering
\small
\begin{tabular}{lrrrrrrrr}
\toprule
Split & N & Orig. Acc. & Final Acc. & $\Delta$ & Corrected & Semantic Fix & Spatial Fix & Verified \\
\midrule
Overall & 7484 & 65.18 & 68.05 & +2.87 & 275 & 86 & 189 & 2049 \\
Easy & 3669 & 72.36 & 74.60 & +2.23 & 119 & 55 & 64 & 791 \\
Hard & 3815 & 58.27 & 61.76 & +3.49 & 156 & 31 & 125 & 1258 \\
View-dep. & 2478 & 64.53 & 66.55 & +2.02 & 75 & 24 & 51 & 727 \\
View-indep. & 5006 & 65.50 & 68.80 & +3.30 & 200 & 62 & 138 & 1322 \\
\bottomrule
\end{tabular}
\caption{Analysis of VLM-based verification on Nr3D.
Verified denotes the number of samples processed by the verification module.
Corrected denotes samples whose predictions are changed from incorrect to correct after verification.
For each corrected sample, the VLM additionally identifies whether the correction is primarily attributed to semantic evidence (Semantic Fix) or spatial reasoning (Spatial Fix).
}
\label{tab:vlm_verification_effect}
\end{table*}

In the main paper, we motivate GuideGround by observing that direct multi-view feature aggregation may weaken viewpoint-specific spatial evidence.
To further validate this observation, we conduct an additional analysis comparing conventional multi-view fusion with independent view-wise grounding evaluation.
This analysis investigates whether candidate viewpoints already contain sufficient grounding information before aggregation and provides empirical motivation for our viewpoint-aware verification strategy.

Specifically, we modify MVT-3DVG and MiKASA to perform grounding independently for each candidate viewpoint while keeping the remaining inference pipeline unchanged. 
A prediction is considered correct if any candidate viewpoint identifies the target, providing an oracle upper bound of the 3DVG model assuming perfect viewpoint selection.

As shown in Table~\ref{tab:motivation}, independent per-view grounding consistently outperforms the original multi-view inference strategy, indicating that the correct target is often identifiable from at least one candidate viewpoint before feature aggregation. 
This suggests that the key challenge is not generating multiple viewpoints, but explicitly selecting the most informative viewpoint for spatial reasoning. 
Motivated by this, GuideGround performs independent per-view grounding and identifies the final target through VLM-based hypothesis verification instead of direct multi-view feature aggregation.

\subsection{Analysis of Semantic Understanding}

To better understand the effect of VLM-enhanced semantic understanding, we analyze two complementary metrics in Table~\ref{tab:nr3d_semantic_error}. 
Semantic Accuracy (S. Acc.) measures whether the predicted object belongs to the correct semantic category regardless of the exact instance, reflecting the model's ability to (1) recognize object semantics and (2) understand the semantic target specified in the referring expression. 
Distractor Error (D. Err.) denotes the gap between S. Acc. and the final grounding accuracy, corresponding to cases where the model correctly identifies the target category but fails to localize the correct instance among semantically similar objects.

The proposed semantic representation consistently improves S. Acc., particularly on the Easy split. 
Compared with MiKASA, our grounding model without verification increases S. Acc. from 84.66\% to 86.94\%, and the complete framework further improves it to 88.36\%. 
Since Easy examples contain relatively limited spatial ambiguity, this improvement mainly reflects stronger semantic understanding enabled by VLM-generated object descriptions rather than enhanced spatial reasoning. 
On the Hard split, the grounding model without verification only achieves comparable S. Acc. to MiKASA. 
We attribute this observation to the substantially stronger spatial ambiguity in Hard examples, where incorrect spatial reasoning may prevent semantically correct candidates from being selected. 
Interestingly, the verification module further improves S. Acc. on this split, suggesting that viewpoint-aware verification can recover semantic predictions affected by ambiguous spatial reasoning.

The D. Err. further reveals the remaining challenge after semantic understanding. 
D. Err. measures failures in distinguishing the referred instance after the semantic category has been correctly identified.
A smaller D. Err. indicates that the model is better able to distinguish the correct target from other objects with the same semantic category. 
Our method exhibits lower D. Err. on most evaluation splits, indicating that stronger semantic representations provide a better foundation for instance-level grounding. 
Nevertheless, the Hard split still exhibits substantially larger D. Err. than the Easy split for all methods, suggesting that many remaining failures arise from distinguishing semantically similar objects based on their spatial relationships rather than recognizing object semantics. 
We further compare the View-dep. and View-indep. subsets.
On the View-indep. subset, our method consistently reduces D. Err., indicating that improved semantic representations directly benefit instance discrimination when viewpoint-dependent reasoning is not required.
In contrast, on the View-dep. subset, the grounding model without verification slightly increases D. Err. despite achieving higher S. Acc. This suggests that once object semantics are correctly identified, the remaining errors are dominated by viewpoint-dependent spatial ambiguity rather than semantic confusion. 
After introducing viewpoint-aware verification, D. Err. is reduced while S. Acc. is further improved, demonstrating that the verification module effectively complements semantic understanding by resolving spatial ambiguities that cannot be addressed by semantic representations alone.
This observation further supports our viewpoint-aware verification module, which explicitly reasons over viewpoint-specific grounding hypotheses to resolve such spatial ambiguities.

\subsection{Analysis of VLM-based Verification}

To analyze the effectiveness of VLM-based verification, we examine how it refines grounding hypotheses on Nr3D.
Table~\ref{tab:vlm_verification_effect} reports grounding performance before and after verification, together with the numbers of verified and corrected samples.
A corrected sample denotes a prediction changed from incorrect to correct after verification, while corrected cases are further categorized according to whether the improvement is mainly attributed to semantic evidence (Semantic Fix) or spatial reasoning (Spatial Fix).

VLM-based verification consistently improves grounding performance across all splits.
On the overall split, verification increases the grounding accuracy from 65.18\% to 68.05\%, yielding a 2.87\% absolute improvement by correcting 275 previously incorrect predictions.
These results demonstrate that the VLM verifier effectively refines the viewpoint-specific grounding hypotheses generated by the task-specific grounding model rather than replacing it.
The corrected cases further provide insights into the role of VLM-based verification.
Among the 275 corrected samples on the overall split, 189 are attributed to spatial reasoning, whereas only 86 are attributed to semantic evidence.
These statistics suggest that spatial reasoning plays a dominant role during verification, while semantic evidence mainly serves as complementary support.

The benefit of verification becomes more evident in challenging scenarios.
On the Hard split, verification corrects 156 samples, improving accuracy by 3.49\%. Among these corrections, 125 are attributed to spatial reasoning.
This suggests that verification is particularly useful when the target must be distinguished from same-category distractors using relational evidence.
On the View-dep. split, spatial corrections still account for the majority of improvements (51 out of 75 corrected samples).
This distribution is consistent with the intended role of the verifier, which compares viewpoint-specific grounding hypotheses to resolve spatial relations expressed under a particular reference frame.
Overall, the results support the complementary design of GuideGround, where the task-specific model performs structured grounding and the VLM verifier refines ambiguous hypotheses through high-level spatial reasoning.

\subsection{More Qualitive Results}
\begin{figure*}[t]
\centering
\includegraphics[width=0.95\textwidth]{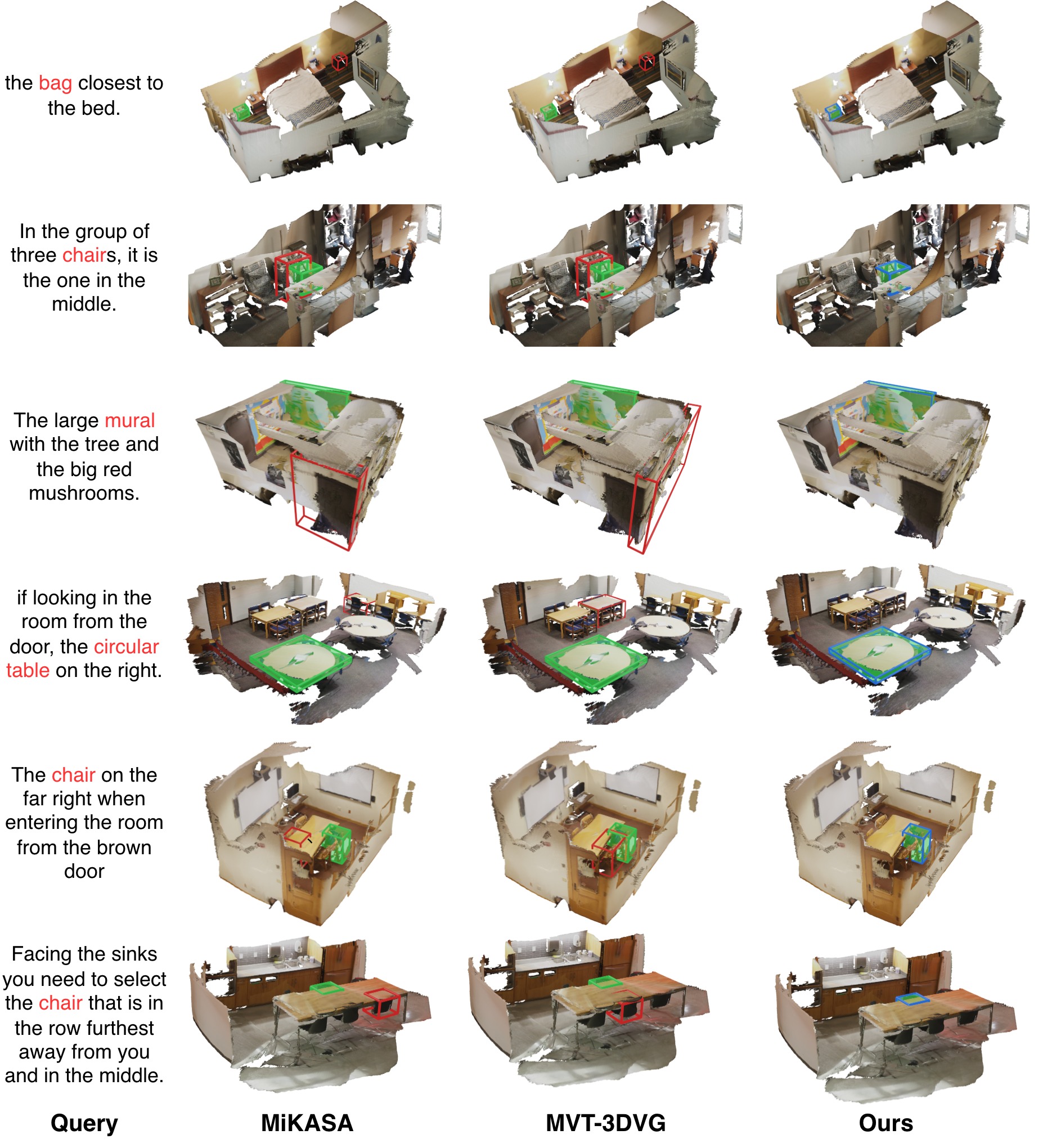}
\caption{
\textbf{Qualitative comparison on challenging 3D visual grounding cases.}
The green,red and blue boxes indicate the ground-truth target, false predictions and the correct predictions respectively.
GuideGround achieves more accurate grounding by leveraging fine-grained semantic descriptions and viewpoint-aware hypothesis verification, especially in cases involving ambiguous object categories and complex spatial relations.
}
\label{fig:qualitative_results}
\end{figure*}

The qualitative results in Figure~\ref{fig:qualitative_results}. demonstrate the effectiveness of GuideGround in challenging grounding scenarios.
Existing methods often fail when multiple objects share similar appearances or when the referring expression requires viewpoint-dependent spatial reasoning.

The first three examples illustrate the challenge of semantic understanding in indoor 3D scenes.
Existing methods may rely heavily on geometric representations or closed-set semantic cues, making them vulnerable to visually similar objects and incomplete observations.
In the first example, the target bag is confused with a semantically related backpack, which shares similar appearance but does not satisfy the query.
In the second example, incomplete object geometry causes the co-located jacket to be interpreted as a chair-like structure, leading to incorrect grounding.
In the third example, the target mural is distinguished by fine-grained visual attributes, such as the depicted tree and red mushrooms, which are difficult to capture through coarse object categories.
By leveraging VLM-generated semantic descriptions, GuideGround captures these object-specific cues and better distinguishes the target from distractors.

The remaining examples demonstrate the difficulty of viewpoint-dependent spatial reasoning.
The target object cannot be identified solely from object categories, but requires interpreting spatial relations under a specific reference viewpoint.
For example, expressions involving "entering the room from the brown door" or "facing the sink" require selecting the viewpoint that matches the described frame of reference.
Direct multi-view aggregation may obscure such viewpoint-specific evidence, whereas GuideGround preserves viewpoint-specific hypotheses and uses VLM-based verification to identify the viewpoint that best explains the query.

%% file: ref.bib
@article{liu2025survey,
  title={A survey on text-guided 3-d visual grounding: Elements, recent advances, and future directions},
  author={Liu, Daizong and Liu, Yang and Huang, Wencan and Hu, Wei},
  journal={IEEE Transactions on Neural Networks and Learning Systems},
  year={2025},
  publisher={IEEE}
}

@inproceedings{chen2020scanrefer,
  title={Scanrefer: 3d object localization in rgb-d scans using natural language},
  author={Chen, Dave Zhenyu and Chang, Angel X and Nie{\ss}ner, Matthias},
  booktitle={European conference on computer vision},
  pages={202--221},
  year={2020},
  organization={Springer}
}

@inproceedings{achlioptas2020referit3d,
  title={Referit3d: Neural listeners for fine-grained 3d object identification in real-world scenes},
  author={Achlioptas, Panos and Abdelreheem, Ahmed and Xia, Fei and Elhoseiny, Mohamed and Guibas, Leonidas},
  booktitle={European conference on computer vision},
  pages={422--440},
  year={2020},
  organization={Springer}
}

@inproceedings{yuan2021instancerefer,
  title={Instancerefer: Cooperative holistic understanding for visual grounding on point clouds through instance multi-level contextual referring},
  author={Yuan, Zhihao and Yan, Xu and Liao, Yinghong and Zhang, Ruimao and Wang, Sheng and Li, Zhen and Cui, Shuguang},
  booktitle={Proceedings of the IEEE/CVF International Conference on Computer Vision},
  pages={1791--1800},
  year={2021}
}

@article{yang2023exploiting,
  title={Exploiting contextual objects and relations for 3d visual grounding},
  author={Yang, Li and Zhang, Ziqi and Qi, Zhongang and Xu, Yan and Liu, Wei and Shan, Ying and Li, Bing and Yang, Weiping and Li, Peng and Wang, Yan and others},
  journal={Advances in Neural Information Processing Systems},
  volume={36},
  pages={49542--49554},
  year={2023}
}

@article{chen2022language,
  title={Language conditioned spatial relation reasoning for 3d object grounding},
  author={Chen, Shizhe and Guhur, Pierre-Louis and Tapaswi, Makarand and Schmid, Cordelia and Laptev, Ivan},
  journal={Advances in neural information processing systems},
  volume={35},
  pages={20522--20535},
  year={2022}
}

@inproceedings{wu2023eda,
  title={Eda: Explicit text-decoupling and dense alignment for 3d visual grounding},
  author={Wu, Yanmin and Cheng, Xinhua and Zhang, Renrui and Cheng, Zesen and Zhang, Jian},
  booktitle={Proceedings of the IEEE/CVF conference on computer vision and pattern recognition},
  pages={19231--19242},
  year={2023}
}

@inproceedings{chang2024mikasa,
  title={Mikasa: Multi-key-anchor \& scene-aware transformer for 3d visual grounding},
  author={Chang, Chun-Peng and Wang, Shaoxiang and Pagani, Alain and Stricker, Didier},
  booktitle={Proceedings of the IEEE/CVF Conference on Computer Vision and Pattern Recognition},
  pages={14131--14140},
  year={2024}
}

@inproceedings{abdelrahman2024cot3dref,
  title={Cot3dref: Chain-of-thoughts data-efficient 3d visual grounding},
  author={Abdelrahman, Eslam and Mohamed, Mohamed Ayman and Ahmed, Mahmoud and Elhoseiny, Mohamed},
  booktitle={International Conference on Learning Representations},
  volume={2024},
  pages={11871--11896},
  year={2024}
}

@inproceedings{he2021transrefer3d,
  title={Transrefer3d: Entity-and-relation aware transformer for fine-grained 3d visual grounding},
  author={He, Dailan and Zhao, Yusheng and Luo, Junyu and Hui, Tianrui and Huang, Shaofei and Zhang, Aixi and Liu, Si},
  booktitle={Proceedings of the 29th ACM international conference on multimedia},
  pages={2344--2352},
  year={2021}
}

@inproceedings{hsu2023ns3d,
  title={Ns3d: Neuro-symbolic grounding of 3d objects and relations},
  author={Hsu, Joy and Mao, Jiayuan and Wu, Jiajun},
  booktitle={Proceedings of the IEEE/CVF Conference on Computer Vision and Pattern Recognition},
  pages={2614--2623},
  year={2023}
}

@inproceedings{roh2022languagerefer,
  title={Languagerefer: Spatial-language model for 3d visual grounding},
  author={Roh, Junha and Desingh, Karthik and Farhadi, Ali and Fox, Dieter},
  booktitle={Conference on Robot Learning},
  pages={1046--1056},
  year={2022},
  organization={PMLR}
}

@inproceedings{huang2022multi,
  title={Multi-view transformer for 3d visual grounding},
  author={Huang, Shijia and Chen, Yilun and Jia, Jiaya and Wang, Liwei},
  booktitle={Proceedings of the IEEE/CVF Conference on Computer Vision and Pattern Recognition},
  pages={15524--15533},
  year={2022}
}

@inproceedings{yang2021sat,
  title={Sat: 2d semantics assisted training for 3d visual grounding},
  author={Yang, Zhengyuan and Zhang, Songyang and Wang, Liwei and Luo, Jiebo},
  booktitle={Proceedings of the IEEE/CVF International Conference on Computer Vision},
  pages={1856--1866},
  year={2021}
}

@article{bakr2022look,
  title={Look around and refer: 2d synthetic semantics knowledge distillation for 3d visual grounding},
  author={Bakr, Eslam and Alsaedy, Yasmeen and Elhoseiny, Mohamed},
  journal={Advances in neural information processing systems},
  volume={35},
  pages={37146--37158},
  year={2022}
}

@inproceedings{zhang2023multi3drefer,
  title={Multi3drefer: Grounding text description to multiple 3d objects},
  author={Zhang, Yiming and Gong, ZeMing and Chang, Angel X},
  booktitle={Proceedings of the IEEE/CVF International Conference on Computer Vision},
  pages={15225--15236},
  year={2023}
}

@inproceedings{radford2021learning,
  title={Learning transferable visual models from natural language supervision},
  author={Radford, Alec and Kim, Jong Wook and Hallacy, Chris and Ramesh, Aditya and Goh, Gabriel and Agarwal, Sandhini and Sastry, Girish and Askell, Amanda and Mishkin, Pamela and Clark, Jack and others},
  booktitle={International conference on machine learning},
  pages={8748--8763},
  year={2021},
  organization={PmLR}
}

@inproceedings{huang2021text,
  title={Text-guided graph neural networks for referring 3d instance segmentation},
  author={Huang, Pin-Hao and Lee, Han-Hung and Chen, Hwann-Tzong and Liu, Tyng-Luh},
  booktitle={Proceedings of the AAAI conference on artificial intelligence},
  volume={35},
  number={2},
  pages={1610--1618},
  year={2021}
}

@inproceedings{guo2023viewrefer,
  title={Viewrefer: Grasp the multi-view knowledge for 3d visual grounding},
  author={Guo, Zoey and Tang, Yiwen and Zhang, Ray and Wang, Dong and Wang, Zhigang and Zhao, Bin and Li, Xuelong},
  booktitle={Proceedings of the IEEE/CVF International Conference on Computer Vision},
  pages={15372--15383},
  year={2023}
}

@inproceedings{shi2024aware,
  title={Aware visual grounding in 3d scenes},
  author={Shi, Xiangxi and Wu, Zhonghua and Lee, Stefan},
  booktitle={Proceedings of the IEEE/CVF Conference on Computer Vision and Pattern Recognition},
  pages={14056--14065},
  year={2024}
}

@inproceedings{huang2025viewsrd,
  title={Viewsrd: 3d visual grounding via structured multi-view decomposition},
  author={Huang, Ronggang and Yang, Haoxin and Cai, Yan and Xu, Xuemiao and Zhang, Huaidong and He, Shengfeng},
  booktitle={Proceedings of the IEEE/CVF International Conference on Computer Vision},
  pages={9726--9736},
  year={2025}
}

@inproceedings{yang2024llm,
  title={Llm-grounder: Open-vocabulary 3d visual grounding with large language model as an agent},
  author={Yang, Jianing and Chen, Xuweiyi and Qian, Shengyi and Madaan, Nikhil and Iyengar, Madhavan and Fouhey, David F and Chai, Joyce},
  booktitle={2024 IEEE International Conference on Robotics and Automation (ICRA)},
  pages={7694--7701},
  year={2024},
  organization={IEEE}
}

@inproceedings{li2025seeground,
  title={Seeground: See and ground for zero-shot open-vocabulary 3d visual grounding},
  author={Li, Rong and Li, Shijie and Kong, Lingdong and Yang, Xulei and Liang, Junwei},
  booktitle={Proceedings of the Computer Vision and Pattern Recognition Conference},
  pages={3707--3717},
  year={2025}
}

@inproceedings{liu2026view,
  title={View-on-graph: Zero-shot 3d visual grounding via vision-language reasoning on scene graphs},
  author={Liu, Yuanyuan and Mei, Haiyang and Zhan, Dongyang and Zhao, Jiayue and Zhou, Dongsheng and Dong, Bo and Yang, Xin},
  booktitle={Proceedings of the AAAI Conference on Artificial Intelligence},
  volume={40},
  number={9},
  pages={7386--7394},
  year={2026}
}

@inproceedings{yuan2024visual,
  title={Visual programming for zero-shot open-vocabulary 3d visual grounding},
  author={Yuan, Zhihao and Ren, Jinke and Feng, Chun-Mei and Zhao, Hengshuang and Cui, Shuguang and Li, Zhen},
  booktitle={Proceedings of the IEEE/CVF Conference on Computer Vision and Pattern Recognition},
  pages={20623--20633},
  year={2024}
}

@article{huang2024chat,
  title={Chat-scene: Bridging 3d scene and large language models with object identifiers},
  author={Huang, Haifeng and Chen, Yilun and Wang, Zehan and Huang, Rongjie and Xu, Runsen and Wang, Tai and Liu, Luping and Cheng, Xize and Zhao, Yang and Pang, Jiangmiao and others},
  journal={Advances in Neural Information Processing Systems},
  volume={37},
  pages={113991--114017},
  year={2024}
}

@misc{jain2025unifying,
      title={Unifying 2D and 3D Vision-Language Understanding}, 
      author={Ayush Jain and Alexander Swerdlow and Yuzhou Wang and Sergio Arnaud and Ada Martin and Alexander Sax and Franziska Meier and Katerina Fragkiadaki},
      year={2025},
      eprint={2503.10745},
      archivePrefix={arXiv},
      primaryClass={cs.CV},
      url={https://arxiv.org/abs/2503.10745}, 
}

@inproceedings{fang2024transcrib3d,
  title={Transcrib3d: 3d referring expression resolution through large language models},
  author={Fang, Jiading and Tan, Xiangshan and Lin, Shengjie and Vasiljevic, Igor and Guizilini, Vitor and Mei, Hongyuan and Ambrus, Rares and Shakhnarovich, Gregory and Walter, Matthew R},
  booktitle={2024 IEEE/RSJ International Conference on Intelligent Robots and Systems (IROS)},
  pages={9737--9744},
  year={2024},
  organization={IEEE}
}

@inproceedings{mcvay2025locate,
  title={LOCATE 3d: Real-world object localization via self-supervised learning in 3D},
  author={McVay, Paul and Arnaud, Sergio and Martin, Ada and Majumdar, Arjun and Jatavallabhula, Krishna Murthy and Thomas, Phillip and Partsey, Ruslan and Dugas, Daniel and Gejji, Abha and Sax, Alexander and others},
  booktitle={Forty-second International Conference on Machine Learning},
  year={2025}
}

@inproceedings{zhu20233d,
  title={3d-vista: Pre-trained transformer for 3d vision and text alignment},
  author={Zhu, Ziyu and Ma, Xiaojian and Chen, Yixin and Deng, Zhidong and Huang, Siyuan and Li, Qing},
  booktitle={Proceedings of the IEEE/CVF International Conference on Computer Vision},
  pages={2911--2921},
  year={2023}
}

@inproceedings{abdelreheem2024scanents3d,
  title={Scanents3d: Exploiting phrase-to-3d-object correspondences for improved visio-linguistic models in 3d scenes},
  author={Abdelreheem, Ahmed and Olszewski, Kyle and Lee, Hsin-Ying and Wonka, Peter and Achlioptas, Panos},
  booktitle={Proceedings of the IEEE/CVF Winter Conference on Applications of Computer Vision},
  pages={3524--3534},
  year={2024}
}

@article{song2020mpnet,
  title={Mpnet: Masked and permuted pre-training for language understanding},
  author={Song, Kaitao and Tan, Xu and Qin, Tao and Lu, Jianfeng and Liu, Tie-Yan},
  journal={Advances in neural information processing systems},
  volume={33},
  pages={16857--16867},
  year={2020}
}

@misc{zhang2025think,
      title={Think Visually, Reason Textually: Vision-Language Synergy in ARC}, 
      author={Beichen Zhang and Yuhang Zang and Xiaoyi Dong and Yuhang Cao and Haodong Duan and Dahua Lin and Jiaqi Wang},
      year={2025},
      eprint={2511.15703},
      archivePrefix={arXiv},
      primaryClass={cs.CV},
      url={https://arxiv.org/abs/2511.15703}, 
}

@inproceedings{lin2025seqvlm,
  title={SeqVLM: Proposal-Guided Multi-View Sequences Reasoning via VLM for Zero-Shot 3D Visual Grounding},
  author={Lin, Jiawen and Bian, Shiran and Zhu, Yihang and Tan, Wenbin and Zhang, Yachao and Xie, Yuan and Qu, Yanyun},
  booktitle={Proceedings of the 33rd ACM International Conference on Multimedia},
  pages={3094--3103},
  year={2025}}

@inproceedings{dai2017scannet,
  title={Scannet: Richly-annotated 3d reconstructions of indoor scenes},
  author={Dai, Angela and Chang, Angel X and Savva, Manolis and Halber, Maciej and Funkhouser, Thomas and Nie{\ss}ner, Matthias},
  booktitle={Proceedings of the IEEE conference on computer vision and pattern recognition},
  pages={5828--5839},
  year={2017}
}

@misc{jose2024dinov2meetstextunified,
  title={DINOv2 Meets Text: A Unified Framework for Image- and Pixel-Level Vision-Language Alignment},
  author={Cijo Jose and Théo Moutakanni and Dahyun Kang and Federico Baldassarre and Timothée Darcet and Hu Xu and Daniel Li and Marc Szafraniec and Michaël Ramamonjisoa and Maxime Oquab and Oriane Siméoni and Huy V. Vo and Patrick Labatut and Piotr Bojanowski},
  journal={arXiv:2412.16334},
  year={2024}
}

@article{oquab2023dinov2,
  title={Dinov2: Learning robust visual features without supervision},
  author={Oquab, Maxime and Darcet, Timoth{\'e}e and Moutakanni, Th{\'e}o and Vo, Huy and Szafraniec, Marc and Khalidov, Vasil and Fernandez, Pierre and Haziza, Daniel and Massa, Francisco and El-Nouby, Alaaeldin and others},
  journal={Transactions on Machine Learning Research Journal},
  year={2024}
}

@misc{loshchilov2017decoupled,
      title={Decoupled Weight Decay Regularization}, 
      author={Ilya Loshchilov and Frank Hutter},
      year={2019},
      eprint={1711.05101},
      archivePrefix={arXiv},
      primaryClass={cs.LG},
      url={https://arxiv.org/abs/1711.05101}, 
}

@misc{liu2024deepseek,
      title={DeepSeek-V3 Technical Report}, 
      author={DeepSeek-AI and Aixin Liu and Bei Feng and Bing Xue and others},
      year={2025},
      eprint={2412.19437},
      archivePrefix={arXiv},
      primaryClass={cs.CL},
      url={https://arxiv.org/abs/2412.19437}, 
}

@misc{qwen3.5,
  title = {{Qwen3.5}: Towards Native Multimodal Agents},
  author = {{Qwen Team}},
  month = {February},
  year = {2026},
  url = {https://qwen.ai/blog?id=qwen3.5}
}

@misc{openai2026gpt56,
  title        = {GPT-5.6: Frontier Intelligence That Scales with Your Ambition},
  author       = {OpenAI},
  year         = {2026},
  howpublished = {\url{https://openai.com/index/gpt-5-6/}}
}
